\documentclass[
]{ceurart}

\usepackage{listings}
\begin{document}

%%
%% Rights management information.
%% CC-BY is default license.
\copyrightyear{2024}
\copyrightclause{Copyright for this paper by its authors.
  Use permitted under Creative Commons License Attribution 4.0
  International (CC BY 4.0).}
\conference{PhysioCHI: Towards Best Practices for Integrating Physiological Signals in HCI, May 11, 2024, Honolulu, HI, USA}

\title{An Adaptive System for Wearable Devices to Detect Stress Using Physiological Signals}

\author[1]{Gelei Xu}
\address[1]{Department of Computer Science and Engineering, University of Notre Dame, U.S.A}
\address[2]{Department of Biomedical
Engineering, Rochester Institute of
Technology, U.S.A}

\author[1]{Ruiyang Qin}
\author[2]{Zhi Zheng}
\author[1]{Yiyu Shi}

\begin{abstract}
  Timely stress detection is crucial for protecting vulnerable groups from long-term detrimental effects by enabling early intervention. Wearable devices, by collecting real-time physiological signals, offer a solution for accurate stress detection accommodating individual differences. This position paper introduces an adaptive framework for personalized stress detection using PPG and EDA signals. Unlike traditional methods that rely on a generalized model, which may suffer performance drops when applied to new users due to domain shifts, this framework aims to provide each user with a personalized model for higher stress detection accuracy. The framework involves three stages: developing a generalized model offline with an initial dataset, adapting the model to the user's unlabeled data, and fine-tuning it with a small set of labeled data obtained through user interaction. This approach not only offers a foundation for mobile applications that provide personalized stress detection and intervention but also has the potential to address a wider range of mental health issues beyond stress detection using physiological signals.
\end{abstract}

%%
%% Keywords. The author(s) should pick words that accurately describe
%% the work being presented. Separate the keywords with commas.

\maketitle
\section{Introduction}

Mental stress is a fundamental element in psychology, medicine, and neuroscience that describes the negative emotions and physiological responses individuals face when encountering difficult situations \cite{giannakakis2019review}. Although stress is a natural response of the organism, certain groups are more vulnerable to its harmful effects \cite{albertetti2020stress}. For instance, children with autism spectrum disorder (ASD) are more likely to experience stress, and their caregivers usually report higher stress levels compared to those with typically developing (TD) children. This parenting stress can adversely impact the mental and physical health of both the caregiver and the family as a whole \cite{yu2023understanding}. Similarly, studies also indicate that outdoor manual laborers, like construction workers, are more prone to feeling stressed than the general population. If this stress is not identified and addressed properly, it can greatly increase the risk of accidents at work \cite{shakerian2021assessing}. To better assist these certain groups, it is essential to monitor their stress levels timely, ideally in real-time.

Wearable technology employs small, affordable, and precise sensors for collecting physiological signals, offering a convenient solution for real-time data collection. Research has established strong connections between physiological signals and stress levels. For example, heart rate variability (HRV) and heart rate (HR), derived from photoplethysmography (PPG) signals, serve as indicators of stress \cite{schiweck2019heart}. Additionally, skin conductance level (SCL) and skin conductance response (SCR) in electrodermal activity (EDA) are linked to both chronic and acute stress \cite{posada2020innovations}.

In stress detection applications using physiological signals, machine learning (ML) models are typically employed to analyze physiological signals and identify stress. For instance, \cite{zhu2022feasibility} compares K-Nearest Neighbor, Logistic Regression, and Random Forests using EDA signals, achieving a maximum accuracy of 85.7\%. While ML provides a straightforward and effective modeling approach in laboratory settings, its application in real-world scenarios is often limited due to several factors. Firstly, physiological data can be challenging to collect, resulting in limited training data from a few individuals. This limitation can lead to models that perform poorly on new users if the training data lacks sufficient population diversity. Additionally, training data is mostly gathered in controlled lab environments, which may not accurately represent the varied settings in which users operate their devices. This discrepancy can cause a domain shift, rendering the model less effective outside the lab.

To address these challenges, we propose to develop a stress detection framework that personalizes models for each individual, aiming for high accuracy across all users by leveraging EDA and PPG signals. Our framework consists of three stages. Initially, we train a generalized model backbone using a 1D-CNN. Subsequently, we train an adapted model for each user using unlabeled physiological signals from wearable devices. Finally, we fine-tune the model with a small set of labeled data obtained through user interactions. We also explore potential concerns and applications of this framework in the Discussion session. Our ultimate goal is to incorporate this framework into a mobile application for initial stress detection, enabling users to monitor their stress levels. Upon detecting stress, we aim to integrate additional intervention strategies to assist users in timely stress relief.

\section{Related Works}

\subsection{PPG and EDA for Stress Indication}

Non-invasive wearable devices with physiological sensors have been widely used to monitor individual's physical conditions and related psychological states. Specifically, PPG and EDA signals are commonly employed for stress detection. PPG signals facilitate the extraction of HR and heart rate variability HRV, indicators of autonomic nervous system (ANS) activity \cite{schiweck2019heart}. The two main components in EDA signals, SCL and SCR, reflect the activity of sweat glands and the sympathetic nervous system (SNS) activity, respectively \cite{posada2020innovations}. Several studies have shown that both the PPG and EDA features can be used to detect stress. For example, \cite{nguyen2022digital} adopted a virtual reality platform together with EDA, PPG, and electrocardiogram (ECG) signals to monitor and reduce frontline healthcare providers’ moral distress. \cite{yu2023understanding} tested the feasibility of understanding stress using EDA and PPG signals for children with ASD and their caregivers. \cite{lima2019heart} used EDA and PPG signals to detect mental stress with Random Forest (RF) and reached an accuracy of 77\%. \cite{zhu2022multimodal} used the EDA signals to classify stress and non-stress conditions using Stacking Ensemble Learning (SCL) and achieved an accuracy of 86.4\%.  

Although these studies have shown promising results in their experiment settings, the real-world applicability of these methods remains uncertain since the traditional ML models are only optimized for the specific dataset and are hard to personalize. For broad-scale applications, although there is a need for models that can be personalized for individual users, collecting a large amount of labeled data for each user is impractical.

\subsection{Domain Adaptation and Personalization}

The performance of deep learning models is constrained by the quantity and variety of labeled data. If the training dataset lacks population diversity, the model may underperform with new users. To enhance model performance on new users' data, it's vital to develop effective strategies. One intuitive approach is to collect labeled data from new users and fine-tune the model accordingly. \cite{jia2022personalized}  introduced a meta-learning based personalization method for patient-specific detection using biosignals. \cite{xu2022achieving} and \cite{chen2020fedhealth} applied fine-tunning and transfer learning, respectively.  However, these methods often require labels from users, which can be time-consuming and inconvenient for the user. Our work relates to domain adversarial training, which learns domain-invariant features without needing user-provided labels. \cite{ganin2016domain} proposed the first domain adversarial training method to tackle the unsupervised domain adaptation problem. \cite{chen2023robust} proposed a siamese optimization of \cite{ganin2016domain} which simplifies the problem of classifying different domains. This approach is promising for addressing data variability in stress detection through biosignals, offering a solution that adapts to new user data seamlessly.
\section{Methodology}

\begin{figure}
    \centering
    \includegraphics[width=1\linewidth]{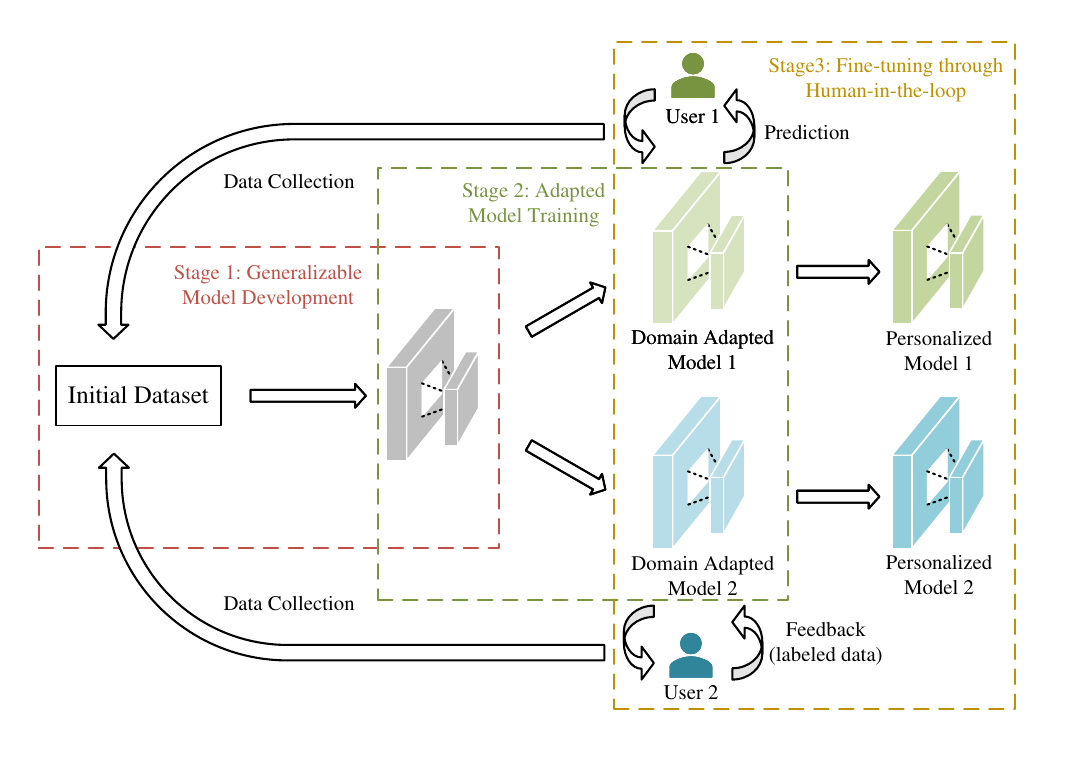}
    \caption{Overview Framework}
    \label{fig:Overview}
\end{figure}

The overall framework is shown in Fig.~\ref{fig:Overview}. This framework is organized into three stages: Initially, it develops a generalized model offline using preliminary labeled data. In the second stage, once the user is registered, the biosensor begins collecting the user's PPG and EDA signals. The model then undergoes domain adaptation by incorporating the user's unlabeled data to become accustomed to their specific signals. Finally, the model is refined with a small set of labeled data collected through user interaction and feedback. This process aims to provide each user with a personalized model that accurately detects stress. Additionally, with user consent for sharing their labeled data, this data can be sent back to enhance the initial dataset, thereby improving the performance of the backbone model in the initial stage.

\subsection{Generalizable Model Development}

\begin{figure}
    \centering
    \includegraphics[width=1\linewidth]{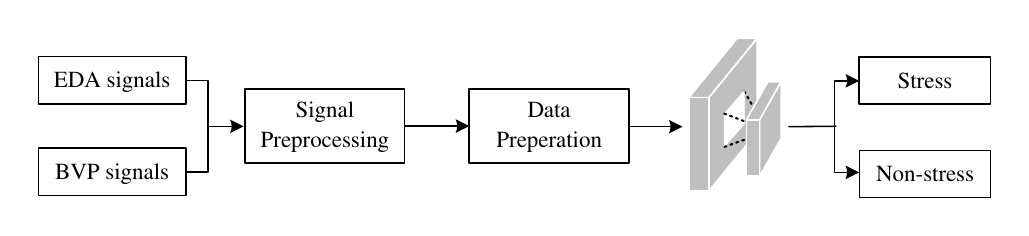}
    \caption{Framework for Training a Generalizable Model}
    \label{fig:Gen}
\end{figure}

The training process for a generalizable model is shown in Fig.~\ref{fig:Gen}. Initially, raw PPG and EDA signals undergo preprocessing to eliminate noise. These signals are then divided into non-overlapping windows, each labeled as stress or non-stress based on user feedback. The final step involves training a CNN model to function as a generalizable model capable of identifying stress levels.

\subsubsection{Preprocessing}

The raw PPG and EDA signals are preprocessed to denoise. For the PPG signal, a band-pass filter (0.5Hz to 5Hz) is applied first to concentrate on relevant frequencies. Then, an adaptive denoising method is applied to remove noise and motion artifacts \cite{heo2021stress}. For the EDA signal, a low-pass filter first reduces noise and motion artifacts, followed by a high-pass filter that distinguishes between SCL (tonic components) and SCR (phasic components) \cite{sanchez2020machine}. The exact preprocessing parameters should be adjusted based on the characteristics of the dataset, such as sampling frequency and the analog prefiltering methods implemented in the hardware.

\subsubsection{Data Preperation}

To facilitate the training of the deep learning model, the preprocessed PPG and EDA signals are segmented into non-overlapping windows of 60 seconds throughout the experiment. 
Since the label is available in the initial dataset at stage 1, each window is labeled as either stress or non-stress, based on the self-reported stress levels of the users.

\subsubsection{Backbone Model}
Several pioneering studies have investigated stress classification through physiological signals, proposing methods like Support Vector Machines \cite{setz2009discriminating}, k-Nearest Neighbour \cite{zhu2022multimodal}, and Random Forest \cite{zhu2022feasibility} to classify stress. Although these methods have demonstrated promising outcomes in specific scenarios, they were optimized for particular datasets with limited subjects. This limitation makes broad-scale applications impractical due to the challenge of collecting extensive labeled data for each user, leading to diminished performance when applied to new users. In contrast, 1D-CNN has been widely used in time-series data classification, excels in capturing the temporal and spatial features of input data and can be easily personalized. Therefore, we believe 1D-CNN with adjustments to its specific architecture based on the input data characteristics is a viable approach for stress detection using physiological signals.

\subsection{Adapted Model Training}

In the first stage, we aimed to develop a generalized model using preliminary labeled data applicable to all users. However, this model, optimized for the training dataset, might underperform for new users, particularly if the training data lacks diversity. A typical solution involves collecting labeled data from the new user to fine-tune the model, but this adds user burden and risks dropout.

We observed that users naturally generate a substantial amount of unlabeled data while using wearable devices. To leverage this, we adopted unsupervised domain adversarial training of neural networks (DANN) \cite{ganin2016domain}, which adapts the generalized model to the new user's data without requiring user-provided labels. DANN employs a feature extractor, a label predictor, and a domain classifier, functioning in a zero-sum game to make the model effective in stress classification without being able to identify the data's user. This approach encourages the extraction of domain-invariant features, enhancing model performance on new user data.

% DANN can learn domain-invariant features from the source domain and target domain. The DANN model consists of three components: a feature extractor, a label predictor, and a domain classifier. The feature extractor is used to extract features from the input data, and the label predictor is used to predict the label of the input data. The domain classifier is used to predict the domain of the input data. The feature extractor and the label predictor are trained to minimize the classification loss on the source domain, while the feature extractor and the domain classifier are trained to maximize the domain confusion on the target domain. By doing so, the feature extractor can learn domain-invariant features that can be used to improve the performance of the model on the target domain.

\begin{figure}
    \centering
    \includegraphics[width=1\linewidth]{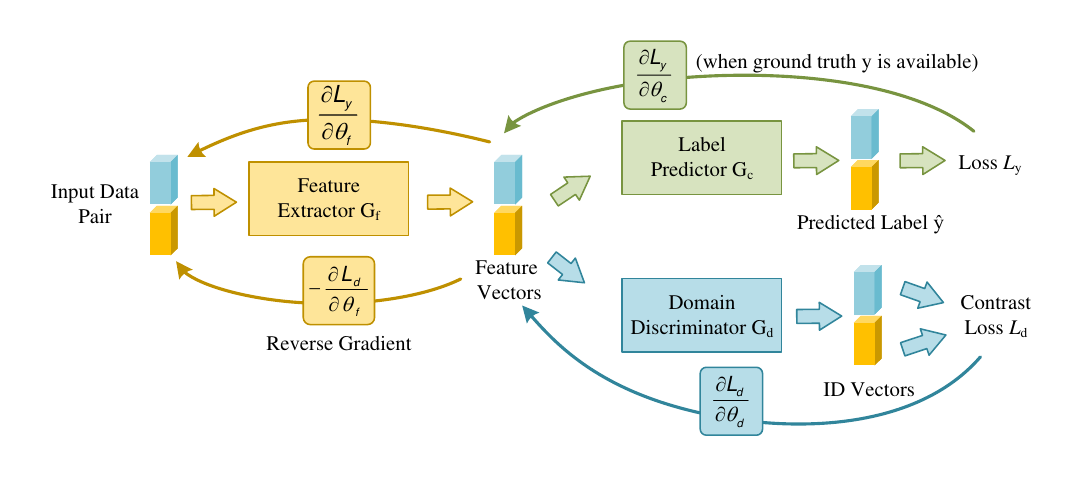}
    \caption{Architecture of the Domain Adversarial Neural Network Model}
    \label{fig:DANN}
\end{figure}

The DANN model's architecture is shown in Fig.~\ref{fig:DANN}. Initially, we randomly select data pairs from both the initial and the target user's datasets. A pair is labeled as positive if both data points are from the same user; otherwise, it is labeled as negative. The concept of data pairing enables the network to solely determine if the input data originates from the current user, eliminating the need to identify all users \cite{chen2023robust}. The input data is processed by the feature extractor $G_f$ to obtain feature representations, which are then used by the label predictor $G_c$ to predict the input data's label. Concurrently, the feature representation is analyzed by the domain discriminator $G_d$ to produce an identification vector for computing the contrastive loss. Ideally, vectors from the same user (positive pairs) should be close, while those from different users (negative pairs) should be separated by a threshold. 
%The final contrastive loss is as follows:

% \begin{equation}
%     L_d = d \cdot \lVert f(x_1) - f(x_2) \rVert_2 + (1 - d) \cdot \max(0, \delta - \lVert f(x_1) - f(x_2) \rVert_2),
% \end{equation}

% where $f(x_1)$ and $f(x_2)$ are the feature representations of the two input data points, $d$ =1 if the two data points are from the same user, and $d$ = 0 otherwise. $\delta$ is a margin that controls the distance between the two feature representations. The contrastive loss is used to train the feature extractor and the domain discriminator to learn domain-invariant features.

After the forward pass, we update the model's parameters by backpropagating the loss. Backpropagation to the label predictor occurs only when the input data's label is known. For the domain discriminator's path, we employ \textit{inverse} backpropagation to reduce the accuracy of domain classification intentionally. This approach ensures targeted learning, optimizing the model's ability to generalize across different domains while maintaining accuracy in label prediction.

\subsection{Fine-tuning through Human-in-the-loop}

During model training with the user's unlabeled data, a small set of labeled data is gathered via user interactions and feedback. This includes confirming stress notifications and actively reporting stress levels.

\textbf{Confirming the Notification.} Users receive a notification when the model predicts stress, asking them to confirm or deny the stress status and adjust the event's timing if necessary. An example of a user interaction prompt is displayed below:

\begin{quote}
    \textit{Are you feeling stressed within the past 5 minutes?}

\textit{[A. Yes, I am stressed, and the timing is from $x$ to $y$ (you may adjust the time)] }

\textit{[B. No, I am not stressed.] }

\textit{[C. I cannot tell.]}
\end{quote}

\textbf{Actively Report Stress Level.} The users are encouraged to report their stress level actively, since the more labeled data the model has, the better the model can assist the user. The timing of these stressful events can also be flexibly adjusted. An example is shown below:

\begin{quote}
\textit{I felt stressed from time $x$ to $y$ / I am feeling stressed now.}
\end{quote}

The labeled data collected from the user is used to refine the model. Incorporating this data into the model aims for better user-specific performance. Additionally, if users consent to share their labeled data, it contributes to improving the overall backbone model's performance in the first stage. 

\section{Discussion}

\subsection{User Data Privacy Concern}

Throughout the process, it is essential to protect user data privacy. An anonymization protocol must be strictly implemented before storing and processing data, removing all personally identifiable information from the datasets. The de-identified data should then be encrypted with cryptographic methods before being sent to backend servers \cite{banerjee2018wearable}. Moreover, users must be clearly informed about the data collection and processing methods, and their consent must be obtained beforehand. They should also be made aware of their rights concerning their data, including access, correction, and deletion rights.

\subsection{On-device Training Constraint}

Given the limited storage and computational resources on a user's device, the feasibility of deploying the model trained in stage 2 directly on the device requires further investigation. Deploying a model trained on backend servers to the user's device could consume significant energy and network resources due to the need for periodic updates \cite{jia2020personalized}. Conversely, training the model on the user's device might be computationally intensive and potentially degrade the user experience. It's crucial to assess the deployment ability of the stage 2 model on a case-by-case basis to balance these factors. Additionally, employing meta-learning \cite{jia2022personalized}, few-shot learning \cite{liu2021few}, and data selection \cite{qin2023enabling} could minimize the training's computational demands.

\subsection{Beyond Stress Detection}
Once stress detection is possible, the system aims to engage with users to pinpoint stress-inducing activities by analyzing the activity patterns of the users. For instance, the system could suggest specific stress-relieving activities like engaging in a hobby, practicing mindfulness, or exercising based on the user's preferences and past responses.

Expanding its utility, the framework has the potential to address a broader spectrum of mental health issues. By leveraging physiological signals, it could adapt to recognize signs of anxiety, depression, and other conditions, offering a versatile tool for mental health monitoring and intervention. This adaptability underscores the system's potential as a comprehensive health management solution, emphasizing its importance in proactive mental wellness.

\section{Conclusion}

This paper introduces a personalized stress detection framework using physiological signals, structured in three stages: developing a generalized model offline with initial labeled data, adapting this model to the user's data, and fine-tuning it with user-provided labeled data. This process aims to create a customized model for each user, capable of accurately detecting stress. The framework is envisioned as a foundation for mobile applications offering personalized stress management solutions. Challenges such as data privacy, on-device training limitations, and the potential for broader mental health applications are also explored, setting the stage for future developments in this area.

\bibliography{ref}

\end{document}